\documentclass[twoside,11pt]{article}

\usepackage[preprint]{jmlr2e}

\usepackage{xcolor}
\usepackage{xurl}
\usepackage{blindtext}

\usepackage{booktabs}
\usepackage{caption}

\usepackage{lastpage}
\jmlrheading{23}{2026}{1-\pageref{LastPage}}{3/26; Revised 3/26}{3/26}{21-0000}
{Charlie F. Ruan, Yucheng Qin, Akaash R. Parthasarathy, Xun Zhou, Ruihang Lai,
Hongyi Jin, Yixin Dong, Bohan Hou, Meng-Shiun Yu, Yiyan Zhai, Sudeep Agarwal,
Hangrui Cao, Siyuan Feng, and Tianqi Chen}

\ShortHeadings{WebLLM: A High-Performance In-Browser LLM Inference Engine}{Ruan et al.}
\firstpageno{1}

\begin{document}

\title{WebLLM: A High-Performance In-Browser LLM \\ Inference Engine}


\author{Charlie F. Ruan\textsuperscript{1},
Yucheng Qin\textsuperscript{1},
Akaash R. Parthasarathy\textsuperscript{1},
Xun Zhou\textsuperscript{1},\\
Ruihang Lai\textsuperscript{1},
Hongyi Jin\textsuperscript{1},
Yixin Dong\textsuperscript{1},
Bohan Hou\textsuperscript{1},
Meng-Shiun Yu\textsuperscript{1},
Yiyan Zhai\textsuperscript{1},
Sudeep Agarwal\textsuperscript{1},
Hangrui Cao\textsuperscript{1},
Siyuan Feng\textsuperscript{2},
Tianqi Chen\textsuperscript{1,3} \\
{\sc Correspondence to: Akaash R.~Parthasarathy, akaashrp@cmu.edu} \\[4pt]
{\it
\textsuperscript{1}Carnegie~Mellon~University, Pittsburgh,~PA, USA, 
\textsuperscript{2}Shanghai~Jiao~Tong~University,\\ Shanghai, China, 
\textsuperscript{3}NVIDIA
}
}


\editor{placeholder}

\maketitle

\begin{abstract}
Advances in large language models (LLMs) have unlocked remarkable capabilities. While deploying these models typically requires server-grade GPUs and cloud-based inference, the recent emergence of smaller open-source models and increasingly powerful consumer devices have made on-device deployment practical. The web browser as a platform for on-device deployment is universally accessible, provides a natural agentic environment, and conveniently abstracts out the different backends from diverse device vendors. To address this opportunity, we introduce WebLLM, an open-source JavaScript framework that enables high-performance LLM inference entirely within web browsers. WebLLM provides an OpenAI-style API for seamless integration into web applications, and leverages WebGPU for efficient local GPU acceleration and WebAssembly for performant CPU computation. With machine learning compilers MLC-LLM and Apache TVM, WebLLM leverages optimized WebGPU kernels, overcoming the absence of performant WebGPU kernel libraries. Evaluations show that WebLLM can retain up to 80\% native performance on the same device with room to close the gap further. WebLLM paves the way for universally accessible, privacy-preserving, personalized, and locally powered LLM applications in web browsers. The code is available at: \url{https://github.com/mlc-ai/web-llm}.



\end{abstract}


\section{Introduction}


Advances in large language models (LLMs) have enabled applications ranging from question-answering and code generation (\cite{roziere2023code}) to more reliable tool use and reasoning-style inference (\cite{openai-o1}, \cite{QwQ}). While frontier models still typically require server-grade accelerators and are hosted on the cloud during inference, on-device use cases have strengthened substantially. Open-weight providers now routinely ship capable small models in the 1-8B-parameter range, and techniques such as quantization have made real-time local inference common on consumer hardware (\cite{abdin2024phi}, \cite{team2024gemma}, \cite{liu2026ministral3}). In parallel, consumer devices have become increasingly suitable for AI workloads. For instance, a 4-bit-quantized 3B-parameter model generates $\sim$90 tokens/s on an Apple M3 laptop (Table~\ref{tab:performance}). Modern laptop-class NPUs are also explicitly marketed around running multi-billion-parameter LLMs locally (\cite{amd2025ryzen}). Together, these trends make on-device LLM deployment both promising and practical. Local inference improves privacy and latency, enables personalization with local data, and unlocks split-execution patterns between co-existing cloud-based and on-device deployments (\cite{qualcomm2023hybrid}).

The web browser is an appealing platform for on-device deployment for three reasons. First, it is a natural agentic environment (\cite{zhou2023webarena}) for tasks such as managing calendars, responding to emails, and creating documents. Second, it is universally accessible: users need only open a URL without installing additional software. Third, it abstracts away the complexities of different device backends. Although mobile devices come from various vendors, browser technologies such as WebGPU are backend-agnostic. Instead of implementing GPU kernels for each backend (e.g.\ CUDA and Metal) for every new LLM operator, developers need only provide a single implementation in WebGPU (\cite{kenwright2022webgpu}) (\S\ref{sec:gpu}).

To address these opportunities, we introduce WebLLM, a high-performance in-browser LLM inference engine. WebLLM is an open-source JavaScript framework that deploys LLMs locally in the browser, providing a practical developer-facing software artifact for integrating LLM capabilities into web applications. It offers an OpenAI-style API for easy integration, leverages WebGPU and machine learning compilers (\cite{chen2018tvm}, \cite{mlc-llm}) for efficient GPU acceleration, uses WebAssembly for performant CPU computation, and supports web workers to separate backend execution and prevent disruption to UI flow. Our evaluation demonstrates that WebLLM retains up to 80\% of the decoding throughput of MLC-LLM on the same device, with potential to further close the gap (\S\ref{sec:eval}).

\section{System Architecture and Key Components}

\begin{figure}[!t]
    \centering
    \includegraphics[width=1\textwidth]{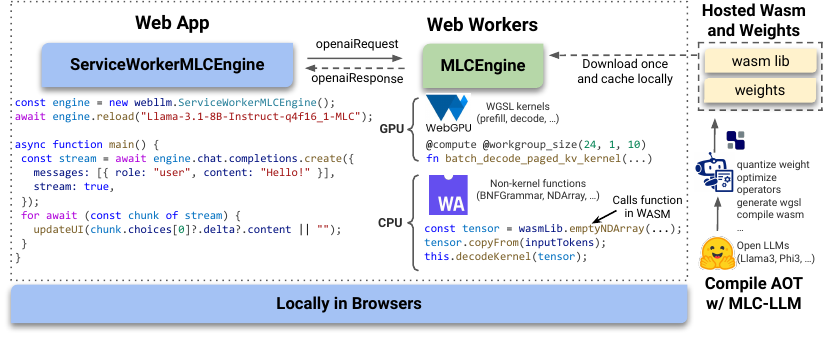}
    \caption{WebLLM System Overview.}
    \label{fig:overview}
\end{figure}

WebLLM is a JavaScript framework that deploys LLMs client-side in the browser, enabling LLM-based features in web applications. Achieving this goal poses three challenges: WebLLM needs (1) a standardized API that web applications can easily incorporate, (2) adaptation to the browser's runtime environment, and (3) efficient GPU acceleration. As shown in Figure~\ref{fig:overview}, WebLLM's architecture addresses these challenges by dividing the system into three corresponding parts: a user-facing engine \texttt{ServiceWorkerMLCEngine} with endpoint-like behavior, an encapsulated \texttt{MLCEngine} that resides in the web worker (a background thread in JavaScript), and ahead-of-time compiled efficient WebGPU kernels.


\subsection{An LLM Inference Engine}

Web developers instantiate a lightweight \texttt{ServiceWorkerMLCEngine} in the web application frontend and treat it like an endpoint. The engine loads an LLM when specified, consumes an OpenAI-style request at any time, and streams back output in an OpenAI-style response, which the web application can use to update the frontend.

This design brings several benefits. Endpoint-like APIs are JSON-in-JSON-out and thus have well-defined behavior. Besides, the OpenAI-style API is widely adopted and makes WebLLM easy to integrate with existing projects. This design also allows WebLLM to enable advanced features with minimal changes to the API. Advanced features that WebLLM supports with this API include: structured generation with JSON Schema and context-free grammar (\cite{dong2024xgrammar}), image input with vision-language models (\cite{abdin2024phi}, \cite{hui2024qwen2}), and loading multiple models in the same engine for applications like retrieval-augmented generation (\cite{lewis2020retrieval}).


%

\subsection{Adapting to the Browser Runtime}

Unlike most LLM inference engines, which are either C++- or Python-based, WebLLM is implemented in JavaScript. This unconventional LLM runtime environment requires WebLLM to adapt to the technologies offered in browsers to ensure high performance.

LLM workloads are computationally heavy and can block the UI if run on the main thread. In JavaScript, web workers are used to separate heavy computation into background threads for a smooth UI. Thus, WebLLM leverages web workers by using two engines: a lightweight frontend engine \texttt{ServiceWorkerMLCEngine} that is exposed to the web application, and a backend engine \texttt{MLCEngine} in the worker thread that actually computes the LLM workload (Figure~\ref{fig:overview}). The two engines communicate via message-passing, and the messages are simply OpenAI-style requests and responses.



Performant LLM inference requires GPU acceleration. WebGPU is a JavaScript API that allows web applications to leverage the native device's GPU in the browser (\cite{kenwright2022webgpu}). WebGPU is backend-agnostic: the same WebGPU kernel can run on devices with different GPU vendors, such as Apple laptops with M chips and laptops with NVIDIA GPUs. Therefore, WebLLM leverages WebGPU for any workload in LLM inference that requires GPU. We discuss in detail how such kernels are generated in \S\ref{sec:gpu}.

Using WebGPU is not enough since LLM inference requires non-trivial computation on the CPU. WebAssembly (WASM) is a portable low-level bytecode that can be compiled from C++ code and run in a JavaScript runtime with near-native performance (\cite{haas2017bringing}). Thus, instead of re-implementing CPU workload in JavaScript, WebLLM leverages Emscripten (\cite{zakai2011emscripten}) to compile high-performance subsystems written in C++ into WebAssembly for various CPU workloads in LLM inference, including a grammar engine for structured generation (\cite{dong2024xgrammar}), sequence management in the paged KV-cache (\cite{mlc-llm}), and tensor manipulation for launching kernels (\cite{chen2018tvm}). This enables C++ code reuse for WebLLM without sacrificing performance.

\subsection{GPU acceleration with WebGPU via MLC-LLM}
\label{sec:gpu}

GPU acceleration is crucial for high-performance LLM inference. WebGPU provides a standardized API to leverage GPU in JavaScript and abstracts out devices with different GPU vendors. However, unlike native backends such as CUDA, WebGPU does not have accelerated GPU libraries for common kernels. This makes it challenging to write high-performance customized GPU kernels such as PagedAttention and FlashAttention for WebGPU (\cite{kwon2023paged}, \cite{dao2022flashattention}).

WebLLM resolves this by leveraging machine learning compilation libraries MLC-LLM and Apache TVM to compile performant WebGPU kernels. MLC-LLM ingests any open-source model's implementation in Python, which can use techniques such as the aforementioned PagedAttention and FlashAttention, and compiles the model's computation into the backend of interest (in this case, WebGPU). Besides compiling to the specified target, MLC-LLM provides both graph-level optimizations (e.g. kernel fusion) and operator-level optimizations (e.g. GEMM tiling) to improve kernel performance.

MLC-LLM converts open-source models into two artifacts: converted weights and a WASM library. The WASM library contains both WebGPU kernels and non-kernel functions in WebAssembly. As shown in Figure~\ref{fig:overview}, the models that WebLLM loads in are compiled ahead-of-time and hosted online. 




\section{Evaluation}
\label{sec:eval}

\begin{table}[h]
    \centering
    \begin{tabular}{lccc}
        \toprule
        \textbf{Model} & \textbf{WebLLM} (tok/s) & \textbf{MLC-LLM} (tok/s) & \textbf{Perf. Retained} \\
        \midrule
        Llama-3.1-8B & 41.1 & 57.7 & 71.2\% \\
        Phi-3.5-mini (3.8B) & 71.1 & 89.3 & 79.6\% \\
        \bottomrule
    \end{tabular}
    \caption{Decoding throughput comparison between WebLLM (v0.2.75) and MLC-LLM (commit d23d6f5). WebLLM is evaluated in Chrome Canary 133.0.6870.0 (arm64).}
    \label{tab:performance}
\end{table}

We evaluate WebLLM by comparing it against MLC-LLM on the same Apple MacBook Pro M3 Max. WebLLM executes in the browser using JavaScript, WebAssembly, and WebGPU, whereas MLC-LLM runs natively on the same hardware using a Python and C++ stack with Metal kernels. Table~\ref{tab:performance} reports decoding throughput in tokens/s for 4-bit quantized models. Across the tested models, WebLLM retains up to $\sim$80\% of the decoding throughput of MLC-LLM on the same device. These results indicate that browser-based deployment can approach the performance of a native deployment while preserving the accessibility and portability advantages of the web platform.

\section{Conclusion}

WebLLM demonstrates that high-performance, on-device LLM inference is feasible directly within the browser. As an open-source JavaScript framework, it empowers developers to integrate advanced LLM capabilities into web applications without sacrificing privacy or requiring server-grade resources. WebLLM thus paves the way for accessible, personalized, and private LLM-driven experiences in everyday web usage.

\bibliography{webllm}

\end{document}